\documentclass[twoside,11pt,preprint]{article}  

\usepackage{blindtext}

\usepackage{graphicx}
\usepackage{jmlr2e}
\usepackage{listings}
\usepackage{wrapstuff}
\usepackage{xcolor}
\usepackage{booktabs}

\DeclareFixedFont{\ttb}{T1}{txtt}{bx}{n}{10}
\DeclareFixedFont{\ttm}{T1}{txtt}{m}{n}{10}

\definecolor{deepblue}{rgb}{0,0,0.5}
\definecolor{deepred}{rgb}{0.6,0,0}
\definecolor{deepgreen}{rgb}{0,0.5,0}

\newcommand*{\revised}[1]{#1}  

\usepackage{lastpage}
\jmlrheading{23}{2025}{1-\pageref{LastPage}}{1/21; Revised 5/22}{9/22}{21-0000}{Yoshihiko Ozaki, Shuhei Watanabe, and Toshihiko Yanase}

\ShortHeadings{OptunaHub: A Platform for Black-Box Optimization}{Ozaki et al.}
\firstpageno{1}

\begin{document}

\title{OptunaHub: A Platform for Black-Box Optimization}

\author{\name Yoshihiko Ozaki\thanks{These authors equally contributed to this work.} \textsuperscript{,1} \email yozaki@preferred.jp \\
        \name Shuhei Watanabe\footnotemark[1] \revised{\textsuperscript{,}\thanks{\revised{This work was conducted while the author was affiliated with Preferred Networks, Inc.}} \textsuperscript{,2} \email shuhei.watanabe@sbintuitions.co.jp} \\
        \name Toshihiko Yanase\textsuperscript{1} \email yanase@preferred.jp \\
        \addr \textsuperscript{1}Preferred Networks, Inc., Otemachi Bldg., 1-6-1 Otemachi, Chiyoda-ku, Tokyo, Japan \\
        \addr \textsuperscript{2}\revised{SB Intuitions Corp., 1-7-1 Kaigan, Minato-ku, Tokyo, Japan}
}

\editor{My editor}

\maketitle

\begin{abstract}
Black-box optimization (BBO) underpins advances in domains such as AutoML and Materials Informatics, yet implementations of algorithms and benchmarks remain fragmented across research communities.
We introduce OptunaHub (\url{https://hub.optuna.org/}), a community-oriented, decentralized platform for distributing BBO components under a unified Optuna-compatible interface.
OptunaHub enables independent publication, discovery, and reuse of optimization algorithms and benchmark problems through a lightweight Python module, a contributor-driven registry, and a searchable web interface.
The source code is publicly available in the \href{https://github.com/optuna/optunahub}{\texttt{optunahub}}, \href{https://github.com/optuna/optunahub-registry}{\texttt{optunahub-registry}}, and \href{https://github.com/optuna/optunahub-web}{\texttt{optunahub-web}} repositories under the Optuna organization on GitHub (\url{https://github.com/optuna/}).
\end{abstract}


\section{Introduction}
\label{sec:intro}

\revised{Black-box optimization (BBO) has enabled significant advances in modern research domains, including AutoML~\citep{hutter2019automated} and Materials Informatics~\citep{terayama2021black}.
These advances have driven the development of increasingly sample-efficient algorithms~\citep{turner2021bayesian} and benchmarking toolkits~\citep{hansen2021coco,doerr2018iohprofiler,eggensperger2021hpobench,pfisterer2022yahpo}.}
\revised{Generally, existing BBO software efforts fall into two categories: 
BBO frameworks such as Optuna~\citep{akiba2019optuna}, Nevergrad~\citep{bennet2021nevergrad}, and pymoo~\citep{blank2020pymoo}, which provide algorithm collections within centrally maintained codebases;
\revised{and benchmarking and profiling toolkits such as COCO~\citep{hansen2021coco} and IOHprofiler~\citep{doerr2018iohprofiler}, which focus primarily on standardized evaluation.}
While these efforts have substantially improved the dissemination of advanced optimization algorithms as well as reproducibility and evaluation practices, they are not intended to distribute modules implemented by independent contributors.
This centralist design promotes coherence and standardization, but algorithm and benchmark implementations often remain fragmented across research communities and software ecosystems, ultimately hindering cross-domain reuse and fast adoption of new optimization methods.}

By comparison, the broader machine learning community has undergone a paradigm shift with the emergence of the Hugging Face Hub, which enables researchers to independently publish, discover, and reuse models and datasets through a unified interface. 
This model has influenced research workflows by improving visibility, interoperability, and reuse~\citep{osborne2024ai}.
Despite the rapid growth of BBO research, a comparable community-driven distribution infrastructure for optimization algorithms, benchmark problems, and related utilities has yet to emerge.

\begin{figure}[t]
    \centering
    \includegraphics[width=\textwidth]{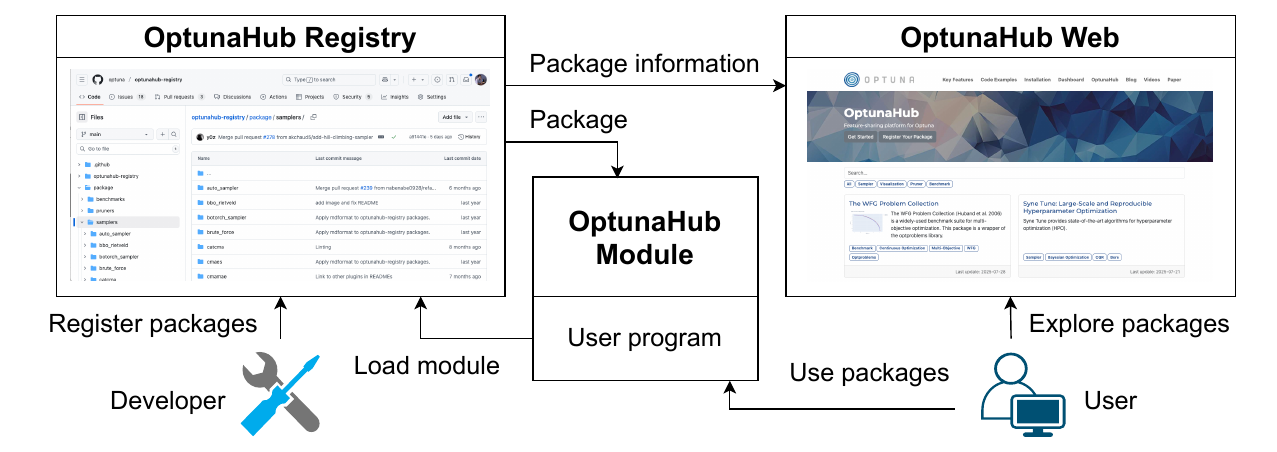}
    \caption{
        Conceptual visualization of the relationships between OptunaHub components.
        The OptunaHub Module enables users to load packages from the OptunaHub Registry, while package information is available via the OptunaHub Web interface.
        All packages integrate seamlessly with the Optuna interface, allowing different algorithms to be applied across problems with minimal modification.
        The Web interface also provides full-text search for efficient package discovery.
    }
    \label{fig:overview}
\end{figure}

\revised{To this end, we propose \emph{OptunaHub} (\url{https://hub.optuna.org/}), a community-oriented platform for BBO.}
OptunaHub consists of (1) a Python library (\emph{OptunaHub Module}) providing unified APIs built on top of Optuna~\citep{akiba2019optuna}, (2) a package registry (\emph{OptunaHub Registry}) that enables contributors to independently develop and distribute algorithms, benchmark problems, and utilities with isolated dependencies, and (3) a web interface (\emph{OptunaHub Web}) that facilitates searchable discovery of registered packages (Figure~\ref{fig:overview}).
\revised{Our goal is to reduce fragmentation in BBO research by fostering cross-domain exchange and sustained interaction between contributors and practitioners. The OptunaHub infrastructure increases not only the visibility and discoverability of optimization algorithms but also their reusability thanks to the unified interface backed by the Optuna assets.
Appendix~\ref{appendix:bbo-software-landscape} further details the position of OptunaHub within the BBO software landscape.}

\section{OptunaHub Ecosystem}

This section describes the three major components of OptunaHub illustrated in Figure~\ref{fig:overview}: OptunaHub Module (a Python library providing unified APIs and utilities for interacting with registry packages), OptunaHub Registry (a repository of contributed packages), and OptunaHub Web (a web interface that aggregates and presents package information). 

\subsection{OptunaHub Module: Unified APIs Compatible with Optuna}
\label{sec:library}

OptunaHub Module, available via \texttt{pip install \href{https://pypi.org/project/optunahub/}{optunahub}}, provides two primary features: (1) the \texttt{load\_module} function and (2) a set of base classes (e.g., \texttt{SimpleBaseSampler} and \texttt{BaseProblem}) for implementing Optuna-compatible interfaces.\footnotemark[1]

\footnotetext[1]{\url{https://optuna.github.io/optunahub/reference.html}}

The \texttt{load\_module} function dynamically imports a module from the OptunaHub Registry package specified as \texttt{load\_module("category/package\_name")}. 
\revised{For example, Lines~\ref{line:load-auto-sampler}–\ref{line:load-nasbench} in Code~\ref{lst:load_module} load \texttt{"samplers/auto\_sampler"}, a meta-sampler that automatically selects an appropriate optimization strategy, and \texttt{"benchmarks/nasbench201"}, a wrapper of NAS-Bench-201~\citep{dong2020nasbench201}.}
Downloaded packages via \texttt{load\_module} are cached locally, enabling subsequent offline use.
Each package has a corresponding web page (Section~\ref{sec:web-interface}) that documents its available APIs.

Through the provided base classes, samplers (i.e., BBO algorithms) and benchmarks conform to the Optuna interface. 
This compatibility enables users to swap samplers and benchmark problems seamlessly and to analyze optimization results using standard Optuna utilities (Lines~\ref{line:visualization}--\ref{line:analysis}).
Optuna itself is a widely adopted optimization framework, and registered packages benefit directly from its mature backend and features.

\begin{lstlisting}[caption={\revised{An example codeblock to run modules via \texttt{optunahub}.}},label={lst:load_module},float=t]
from optuna import create_study
from optuna.visualization import plot_optimization_history, plot_param_importances
from optunahub import load_module

# All modules have their own package pages at https://hub.optuna.org/.
auto_sampler = load_module("samplers/auto_sampler") |\label{line:load-auto-sampler}|
nasbench201 = load_module("benchmarks/nasbench201") |\label{line:load-nasbench}|

auto_sampler = auto_sampler.AutoSampler()
objective = nasbench201.Problem(dataset_id=0)  # Use the CIFAR-10 dataset.
study = create_study(directions=objective.directions)
study.optimize(objective, n_trials=30)
plot_optimization_history(study).show() |\label{line:visualization}|
plot_param_importances(study).show() |\label{line:analysis}|
\end{lstlisting}

\subsection{OptunaHub Registry: Gateway for Community Contributions}
\label{sec:registry}

OptunaHub Registry enables researchers and practitioners to share modular packages of BBO functionality that are accessible through the Optuna interface. 
\revised{Each package---typically an optimization algorithm or benchmark suite---behaves as an independent Python package. These packages can be fetched individually via \texttt{load\_module}, and may specify its own dependencies. Importantly, the fact that OptunaHub accepts new packages from community differentiates from traditional BBO libraries such as Optuna~\citep{akiba2019optuna}, Nevergrad~\citep{bennet2021nevergrad}, and pymoo~\citep{blank2020pymoo}, which distribute a predefined set of functionalities. Appendix~\ref{appendix:registry-and-package-details} details the package registration process, registry structure, and package components.}

As of this writing, more than 100 packages are available.
The sampler collection ranges from classical methods, such as Nelder–Mead~\citep{nelder1965simplex}, to recent approaches including HEBO~\citep{cowen2022hebo}, Vizier~\citep{song2022open}, and LLM-enhanced Bayesian optimization~\citep{liu2024large}. 
Several methods have been contributed by their original authors, including CatCMA~\citep{hamano2024catcma}, SMAC~\citep{lindauer2022smac3}, SyneTune~\citep{salinas2022syne}, and PFNs4BO~\citep{muller2023pfns4bo}.
The registry also accepts packages providing benchmarks, pruning methods, and visualization tools. 
Registered benchmarks span synthetic function suites such as BBOB~\citep{hansen2009real} and WFG~\citep{huband2006review}, AutoML-related tasks including HPO/NAS-Bench~\citep{klein2019tabular,eggensperger2021hpobench,dong2020nasbench201,dong2021nats}, and real-world problem benchmarks such as aircraft design~\citep{namura2025single}. 

\subsection{OptunaHub Web: Package Catalog with Full-Text \& Tag Search}
\label{sec:web-interface}

\begin{wrapstuff}[type=figure,width=0.4\textwidth,top=0]
    \centering
    \includegraphics[width=\textwidth]{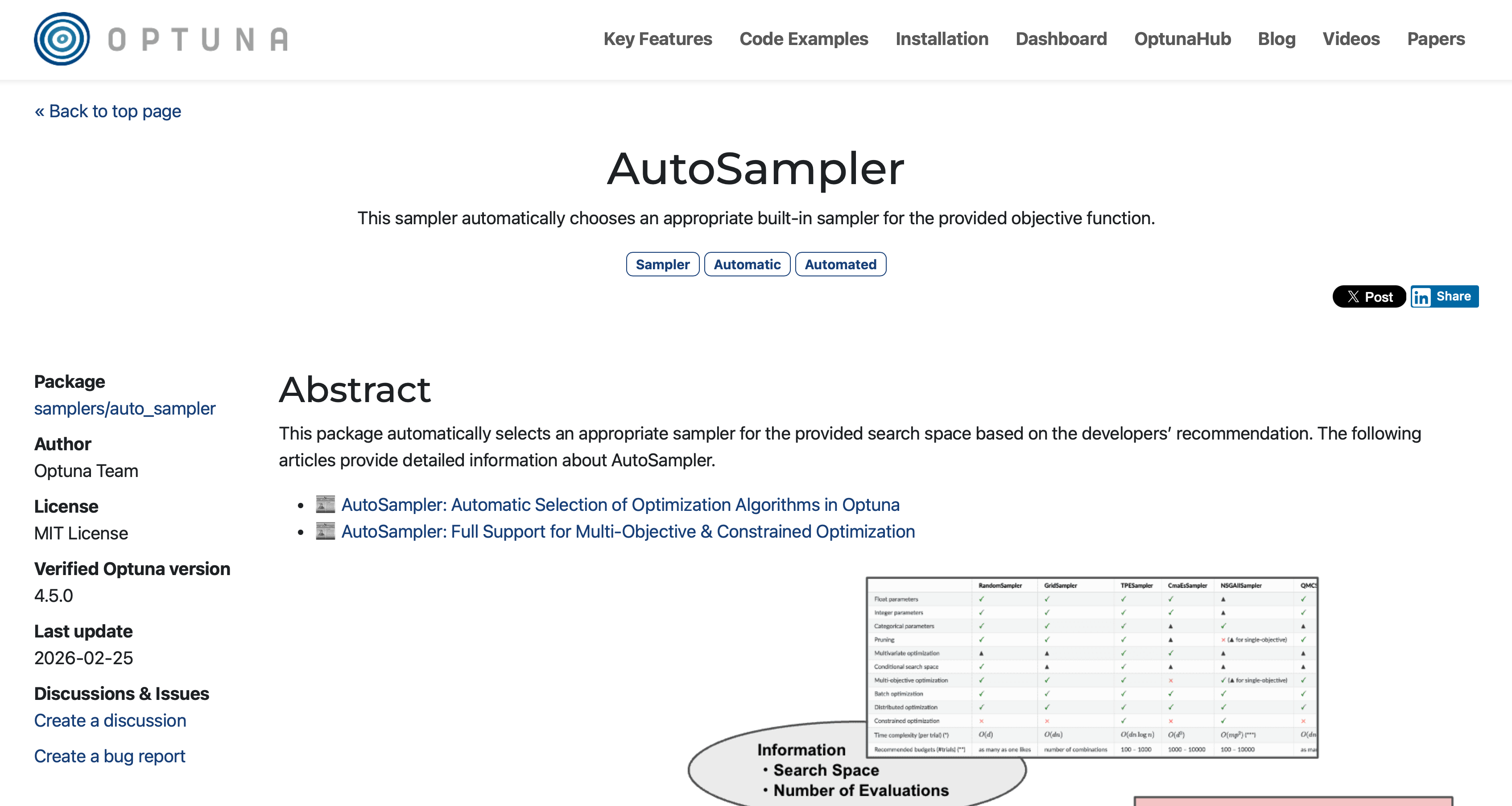}
    \caption{\revised{Package page example.}}
    \label{fig:optunahub-web}
\end{wrapstuff}

OptunaHub Web serves as the discovery and presentation layer of the ecosystem, enhancing package accessibility.
It provides individual package pages that are automatically generated from each package’s \texttt{README.md} (cf.~Appendix~\ref{appendix:registry-and-package-details}), ensuring that documentation remains synchronized with the source repository. 
In addition, the Web interface offers full-text and tag-based search capabilities, enabling users to efficiently locate relevant algorithms, benchmarks, and utilities across domains. 
The top page presents a structured list view of package thumbnails, tags, and summaries to facilitate browsing.
Each package page includes author and license information, a summary, tags, dependency details, API documentations, code examples, \revised{and links for discussion or bug reporting.}
\revised{Figure~\ref{fig:optunahub-web} illustrates the \texttt{auto\_sampler} package page as an example.\footnotemark[2]}
Package pages are automatically refreshed whenever the corresponding \texttt{README.md} file in the OptunaHub Registry is updated, ensuring consistency between registry content and web documentation. 
By standardizing how packages are documented and presented, OptunaHub Web improves discoverability and reuse while providing contributors with a consistent mechanism for disseminating and maintaining their work.

\footnotetext[2]{\revised{\url{https://hub.optuna.org/samplers/auto_sampler/}}}

\section{Final Remarks}

Open-source software and collaborative development play a central role in modern scientific research.
In this paper, we introduced OptunaHub, a decentralized distribution platform for BBO components built on top of Optuna.
By combining modular package development with a unified execution interface, OptunaHub enables independent contribution while preserving interoperability and reproducibility.
As the ecosystem continues to evolve, we expect OptunaHub to complement existing BBO libraries and benchmarking toolkits by providing infrastructure for community-driven dissemination of algorithms and benchmark problems. 
We hope this platform will contribute to broader cross-domain reuse and sustained progress in BBO research.

\acks{\revised{We thank the editor and the anonymous reviewers for their constructive comments and suggestions.} We also thank Kei Akita, Kaito Baba, Yugo Fusawa, Hideaki Imamura, Naoto Mizuno, Masashi Shibata, Hiroki Takizawa, and Yunzhuo Wang for their support. We acknowledge the BBO community, particularly the contributors to OptunaHub. We thank Yasuhiro Fujita for his advice on this submission.}

\bibliography{optunahub}

\appendix

\section{\revised{OptunaHub's Position within the BBO Software Landscape}}
\label{appendix:bbo-software-landscape}

This appendix provides a structured comparison of OptunaHub with representative software efforts in BBO. 
Our goal is not to rank existing tools, but to clarify differences in design philosophy, governance structure, and intended use cases.

\subsection{\revised{Library-Centric BBO Frameworks}}

\revised{Optuna~\citep{akiba2019optuna}, Nevergrad~\citep{bennet2021nevergrad}, pymoo~\citep{blank2020pymoo}, and several recently proposed BBO frameworks~\citep{tian2017platemo,van2023mealpy,jiang2024openbox} provide extensive collections of optimization algorithms (and some of them also provide benchmark problems) within centrally maintained codebases. 
These libraries offer rich functionality and coherent APIs, enabling users to apply a variety of algorithms through a unified interface. 
New features and algorithms are typically integrated directly into a shared codebase and maintained under unified project governance.
Such centralized integration facilitates consistency and quality control. 
However, the primary objective of these frameworks is to provide comprehensive optimization toolkits rather than to function as open distribution platforms for third-party components.}

\revised{Moreover, some libraries are developed with a specific methodological focus or problem domain in mind. 
For example, pymoo and PlatEMO~\citep{tian2017platemo} are primarily oriented toward evolutionary multi-objective optimization within the broader BBO landscape.}

\subsection{\revised{Benchmarking and Profiling Toolkits}}

\revised{Benchmarking and profiling toolkits such as COCO~\citep{hansen2021coco} and IOHprofiler~\citep{doerr2018iohprofiler} focus on standardized evaluation and performance analysis of optimization algorithms.
They provide curated benchmark suites, experimental protocols, and profiling tools to support reproducible comparison. These toolkits advance methodological rigor in BBO research. However, their design centers on carefully curated evaluation infrastructures, where benchmark suites are defined and maintained within a centralized framework rather than distributed as independently developed packages.}

\revised{In terms of problem coverage, the benchmarking toolkits mentioned above emphasize synthetic test functions, such as BBOB~\citep{hansen2009real}, designed to analyze algorithmic properties under controlled conditions. 
Some benchmark collections instead focus on specific application domains, e.g., HPO/NAS-Bench~\citep{klein2019tabular,eggensperger2021hpobench,dong2020nasbench201,dong2021nats} target AutoML tasks.
While these efforts provide valuable resources for their respective purposes, they are typically distributed as curated datasets or evaluation toolkits rather than as open registries enabling third-party contributors to publish and maintain diverse benchmark implementations.}

\revised{OptunaHub complements these approaches by enabling contributors to implement and distribute benchmark problems as modular packages under a unified interface.
This decentralized distribution model allows the ecosystem to expand beyond predefined synthetic suites or domain-specific collections, supporting a broader range of application-driven and community-contributed benchmarks. Importantly, OptunaHub Web enhances searchability across domains and communities.}

\subsection{\revised{OptunaHub’s Design Philosophy}}

\revised{OptunaHub differs from the above projects in its governance and distribution model. Rather than integrating all functionality into a shared codebase, it adopts a decentralized package registry in which each component---such as an optimization algorithm or benchmark problem---can be developed, versioned, and maintained independently by its contributors.
All packages conform to Optuna-compatible interfaces, ensuring runtime interoperability, while implementation details and dependencies remain encapsulated within each package. This modular separation allows contributors to develop components independently without modifying or relying on other packages.
The review process follows lightweight criteria that emphasize basic functionality, safety, and ethical compliance, thereby lowering the contribution barrier while preserving ecosystem integrity (cf. Appendix~\ref{appendix:registry-and-package-details}).
Additionally, OptunaHub Web facilitates the discovery of registered packages and access to their contributor-maintained documentation, enhancing visibility and usability across the ecosystem.}

\revised{Overall, OptunaHub introduces a complementary distribution layer that enables decentralized publication and discovery of BBO components under a unified execution interface, rather than replacing existing libraries or benchmarking toolkits.
Furthermore, OptunaHub is designed as a method-agnostic, domain-flexible infrastructure: it does not prioritize specific optimization paradigms, such as evolutionary algorithms or Bayesian optimization, or predefined problem classes, but instead provides a unified interface through which diverse algorithms and application domains can coexist within a shared ecosystem.}

\section{\revised{OptunaHub Package Registration \& Development}}
\label{appendix:registry-and-package-details}

This appendix describes the package registration process and outlines the structure of the OptunaHub Registry and its packages.

\subsection{\revised{Overview of the Package Registration Process}}

\revised{OptunaHub accepts new package registrations and updates through pull requests to the \href{https://github.com/optuna/optunahub-registry}{\texttt{optunahub-registry}} repository on GitHub. 
Pull requests are reviewed by at least one OptunaHub committer to ensure basic functionality (by verifying a minimal working example), safety (e.g., absence of malicious code), compliance with licensing requirements, and adherence to general research and engineering ethics. 
Each package is primarily maintained by its contributors, i.e., OptunaHub committers do not assume responsibility for individual package implementation details or long-term maintenance.
This separation of responsibilities clarifies that package development is contributor-driven, while the registry structure and submission standards are managed by the OptunaHub committers.
Once these minimal criteria are satisfied, contributions are merged in a timely manner.}

\sloppypar{\revised{OptunaHub's review practices differ from the stricter quality assurance standards adopted by Optuna.
Rather than requiring extensive design reviews or comprehensive unit testing, OptunaHub follows lightweight review criteria that emphasize basic functionality and safety.
This approach balances baseline quality with broad community participation.}}

\subsection{Registry \& Package Details}

OptunaHub Registry follows a category--package directory hierarchy, where each package resides in its corresponding category.
\revised{An OptunaHub package is a self-contained directory, similar to a standard Python package, that implements specific functionality with its own dependencies.
While all packages conform to the Optuna-compatible interface, contributors can be agnostic to other packages and focus on developing their packages independently. 
As described in Section~\ref{sec:library}, packages can be loaded directly in user code via \texttt{load\_module("category/package\_name")} without \texttt{pip install} or explicit \texttt{import}.
OptunaHub package versions correspond to Git commits. By specifying \texttt{ref="commit\_hash"} in \texttt{load\_module}, users can load the version corresponding to a particular commit of \href{https://github.com/optuna/optunahub-registry}{\texttt{optunahub-registry}}.
If no reference is provided, the latest version is loaded by default.}

\begin{wrapstuff}[type=figure,width=0.35\textwidth,top=0]
    \centering
    \includegraphics[width=\textwidth]{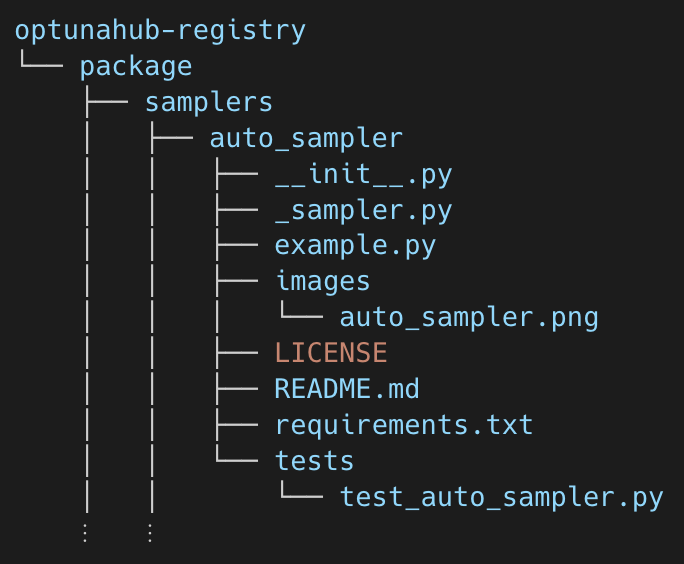}
    \caption{A package example.}
    \label{fig:package}
\end{wrapstuff}

Figure~\ref{fig:package} shows the structure of the \texttt{auto\_sampler} package as an example of a package.
The package is located in the ``samplers'' directory as it belongs to the ``samplers'' category. In the package, \texttt{\_\_init\_\_.py} and \texttt{\_sampler.py} are the implementation of the functionality. \texttt{LICENSE} is the software license file for the package, which must be included. \texttt{README.md} contains metadata (e.g., abstract, authors, and tags) and descriptions of the package (typically, API documentations, benchmark results, and citations), which is published as a package detail page (cf. Section~\ref{sec:web-interface}).
Image files used in \texttt{README.md} can be placed in the ``images'' directory (the first appearing image is used as the thumbnail of the package).
\texttt{requirements.txt}, which is optional, lists the package-specific dependencies.
\revised{Finally, contributors are encouraged to include test code in their packages; however, to lower the barrier to contribution, unit tests are not strictly required at the time of initial submission. 
For review purposes, a functional example demonstrating basic usage must be provided either in the pull request description, in \texttt{README.md}, or in a separate file (e.g., \texttt{example.py}).}
Further instructions are available in the official tutorials for contributors.\footnotemark[3]

\footnotetext[3]{\url{https://optuna.github.io/optunahub/tutorials_for_contributors.html}}

\end{document}